\documentclass[]{spie}  %>>> use for US letter paper
\usepackage{multicol}
\usepackage{fixltx2e} 
\usepackage{amsmath,amsfonts,amssymb}
\usepackage{graphicx,float}
\usepackage{gensymb}
\usepackage{booktabs}       % professional-quality tables
\usepackage[colorlinks=true, allcolors=blue]{hyperref}
\usepackage[font={sf}]{caption}
\title{Physical Reservoir Computing with Origami and its Application to Robotic Crawling}
\author{Priyanka Bhovad}
\author{Suyi Li}
\affil{Department of Mechanical Engineering, Clemson University, Clemson, SC, US}
\authorinfo{Further author information: (Send correspondence to P.B.)\\P.B.: E-mail: pbhovad@clemson.edu\\  S.L.: E-mail: suyil@clemson.edu}

\begin{document} 
\maketitle

\begin{abstract}
A new paradigm called physical reservoir computing has recently emerged, where the nonlinear dynamics of high-dimensional and fixed physical systems are harnessed as a computational resource to achieve complex tasks.  Via extensive simulations based on a dynamic truss-frame model, this study shows that an origami structure can perform as a dynamic reservoir with sufficient computing power to emulate high-order nonlinear systems, generate stable limit cycles, and modulate outputs according to dynamic inputs.  This study also uncovers the linkages between the origami reservoir's physical designs and its computing power,  offering a guideline to optimize the computing performance.   Comprehensive parametric studies show that selecting optimal feedback crease distribution and fine-tuning the underlying origami folding designs are the most effective approach to improve computing performance.  Furthermore, this study shows how origami's physical reservoir computing power can apply to soft robotic control problems by a case study of earthworm-like peristaltic crawling without traditional controllers.   These results can pave the way for origami-based robots with \textit{embodied mechanical intelligence}. 
\end{abstract}


\newpage

% \begin{multicols}{2}
\section{INTRODUCTION} \label{sec:intro} 
The animal kingdom is an endless source of inspiration for soft robotics \cite{Miriyev2020, Laschi2016}.  Researchers have constructed compliant robots that can mimic all kinds of animal motions, like octopus locomotion \cite{Cianchetti2015}, elephant trunk grasping \cite{Hannan2003}, insect flying \cite{Ma2013}, jellyfish and fish swimming \cite{Joshi2019, Ren2019, Katzschmann2016}, as well as snake and insects crawling \cite{Rafsanjani2018, Wu2019, Seok2013}.  These robots share many similarities with animals regarding their shape and motion kinematics; however, their underlying sensing, actuation, and control architectures could be fundamentally different.  Our engineered soft robots typically rely on a centralized controller (aka. an ``electronic brain'') that takes up all computing work to process sensor information, generate control commands, and make decisions.  This approach often struggles to achieve high actuation speed and control effectiveness as soft robots exhibit virtually infinite degrees of freedom and complicated dynamic characteristics.  On the other hand,  animals have highly interconnected networks of nerves and muscles that can share the workload with the brain \cite{Hoffmann2017, Laschi2016b, Trivedi2008}.  The animal body's morphology is an integral part of its actuation, control, and ultimately its ``brain's'' decision-making process, leading to far superior efficiency than our engineered soft robots.

Motivated by this disparity, an increasing number of researchers have embraced soft bodies' nonlinear dynamics as a computational resource to create an embodied intelligence and control \cite{Paul2004, Paul2006a, Caluwaerts2013, Hauser2011, Nakajima2013, Muller2017, Tanaka2019}. As a result, a new computational paradigm called \textit{morphological computation} has emerged in which the physical body of the robot itself takes part in performing low-level control tasks, such as locomotion coordination and modulation, to simplify the overall control architecture significantly \cite{Paul2004, Paul2006a, Hauser2011, Caluwaerts2013, Fuchslin2013}.  The contributions of body morphology to cognition and control involve three major categories \cite{Muller2017}: (1) Morphology facilitating control: wherein the physical design enables certain behaviors such as motion sequencing (e.g., passive dynamic walker \cite{Collins2001}). (2) Morphology facilitating perception: wherein the physical design enables sensing (e.g., the nonuniform distribution of cells in the compound eyes of fly \cite{Floreano2013}). (3) Morphological computation, such as the \emph{physical reservoir computing} (PRC), wherein a physical body performs genuine computations.  Among these, physical reservoir computing shows promising potentials because of its balanced simplicity and versatility to perform applicable computation with encoding and decoding \cite{Muller2017}.

Reservoir computing is a computational framework based on artificial recurrent neural networks (RNNs), which have been used extensively for problems involving time-series prediction like the stock market and weather forecasting, robotic motion planning and control, text and speech recognition \cite{Jaeger2001, Maass2002, Maass2007, Maass2011, Lukosevicius2009, Schrauwen2007, Tanaka2019, Nakajima2020}.  In RNNs, the output of the current time step depends on the results from the previous time step in addition to the current input.   Since RNNs involve both forward and back-propagation of input data, training them became a challenging task.   To address this difficulty, Jaeger introduced the concept of a \emph{fixed} recurrent neural network as Echo State Networks (ESNs) \cite{Jaeger2001}, and Maass introduced Liquid State Machines (LSMs) \cite{Maass2002}.  Later, these two concepts merged under the umbrella of reservoir computing (RC).   In RC, the neural network (aka. the ``reservoir'') has fixed interconnections and input weights, and only the linear output readout weights are trained by simple techniques like linear or ridge regression.  These reservoirs' dynamics transform the input data stream into a high-dimensional state space, capturing its non-linearities and time-dependent information for computation tasks.  

More importantly, the reservoir's fixed nature opens up the possibility of using physical bodies --- such as a random network of nonlinear spring and mass oscillators \cite{Hauser2011, Hauser2012, Morales2018}, tensegrity structures \cite{Paul2004, Paul2006a, Caluwaerts2011, Caluwaerts2013}, and soft robotic arms \cite{Nakajima2013, Li2012, Nakajima2018} ---  to conduct computation, hence the paradigm of \textit{Physical} Reservoir Computing.  These physical systems have shown to possess sufficient computational power to achieve complex computing tasks like emulating other non-linear dynamic systems, pattern generation \cite{Caluwaerts2011, Caluwaerts2013, Hauser2011, Hauser2012, Nakajima2013, Tanaka2019},  speech recognition \cite{Fernando2003}, and machine learning \cite{Nakajima2018, Tanaka2019, Nakajima2020, Morales2018}.  More importantly, robotic bodies with sufficient nonlinear dynamics can also perform like a physical reservoir and directly generate locomotion gait without using the traditional controllers \cite{Caluwaerts2013, Tanaka2019, Degrave2015, Agogino2013, Urbain2017}.

In this study, we investigate the use of origami as a physical reservoir and harness its computing power for robotic locomotion generation.  Origami is an ancient art of folding paper into sophisticated and three-dimensional shapes.  Over the past decades, it has evolved into an engineering framework for constructing deployable structures \cite{Filipov2016, Morgan2016, Dang2020}, advanced materials \cite{Schenk2013, Silverberg2014, Yasuda2019, Li2019, Yan2016, Kamrava2017}, and robotics \cite{Rus2018, Miyashita2017, Belke2017, Onal2015, Onal2013, Yan2018, Novelino2020}.  Origami has many appealing advantages for use in robotics.  It is compact, easy to fabricate, and scale-independent (aka. Origami robots can be fabricated at different scales but still follow similar folding principles \cite{Peraza2014, Rus2018, Ning2018, Morris2016}).  Moreover, the nonlinear mechanics and dynamics induced by folding could enhance robotic performance \cite{Bhovad2019, Zhakypov2015}.  

We show that origami's nonlinear folding dynamics also possess significant computing power.  A mechanical system must exhibit several essential properties to perform as a reservoir \cite{Tanaka2019}.  The first one is high-dimensionality, which allows the reservoir to gather as much information possible from the input data stream, separating its spatio-temporal dependencies and projecting it onto a high-dimensional state-space.  The second one is non-linearity so that the reservoir acts as a nonlinear filter to map the information from the input stream.  All the computation complexity is associated with this nonlinear mapping, thus training the linear static readout becomes a straightforward task.  The third one is fading memory (or short-term memory), ensuring that only the recent input history influences the current output.  The fourth one is separation property to classify and segregate different response signals correctly, even with small disturbances or fluctuations.  Moreover, if two input time series differed in the past, the reservoir should produce different states at subsequent time points \cite{Legenstein2019}.  Our physics-informed numerical simulations prove that origami inherently satisfies these four requirements and can complete computation tasks like emulation, pattern generation, and output modulation.

Moreover, we conduct extensive numerical simulations to uncover the linkage between origami design and its computing power, providing the guideline to optimize computing performance.  Finally, we demonstrate how to directly embed reservoir computing in an origami robotic body to generate earthworm-like peristalsis crawling without using any traditional controllers.  This study's results could foster a new family of origami-based soft robots that operate with simple mechatronics, interact with the environment through distributed sensor and actuator networks, and respond to external disturbances by modulating their activities.

In what follows: Section (\ref{sec:Orig_PRC}) details the construction of an origami reservoir, including the lattice framework used to simulate its nonlinear dynamics. Section \ref{sec:Res} elucidates the origami reservoir's computing power through various numerical experiments.  Section \ref{sec:Para} discusses the parametric analysis that uncovers the linkages between computing performance and physical design.  Section \ref{sec:soro} applies the reservoir computing to an origami robot's crawling problem.  Finally, Section \ref{sec:end} concludes this paper with a summary and discussion.

\section{Constructing The Origami Reservoir} \label{sec:Orig_PRC}
In this study, we construct a physical reservoir using the classical Miura-ori sheets.  It is essentially a periodic tessellation of unit cells, each consisting of four identical quadrilateral \emph{facets} with \emph{crease} lengths $a$ $b$ and an internal sector angle $\gamma$ (Figure \ref{fig:truss} (a)) \cite{Schenk2011,Schenk2013}. The folded geometry of Miura-ori can be fully defined with a dihedral \emph{folding angle} $\theta$ ($\in {[-\pi/2,\pi/2]}$) between the $x$-$y$ reference plane and its facets. The reservoir size is defined as $n \times m$, where $n$ and $m$ are the number of origami \emph{nodes} (aka. vertices where crease lines meet) in $x$ and $y$-directions, respectively.  $N$ is the total number of creases in the origami reservoir. 

\subsection{Dynamics Modeling of the Origami}

\begin{figure}[t]
%	\centering
	\includegraphics[scale=1.0]{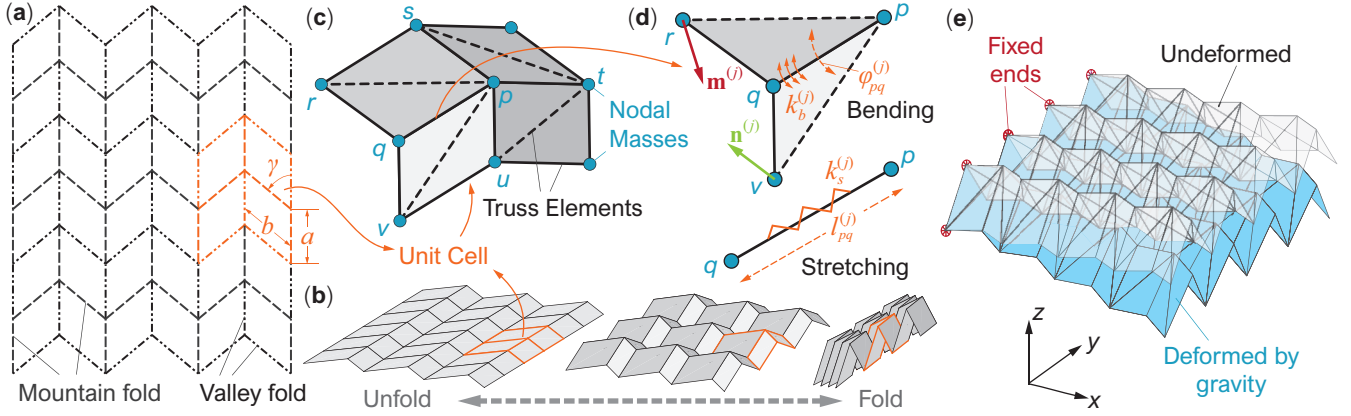}
	\vspace{0.1in}
	\caption{The nonlinear Truss-frame approach for simulating the origami dynamics. (a) The crease pattern of the classical Miura-ori, with a unit cell highlighted.  (b) The rigid-folding kinematics of the Miura-ori. (c) The truss-frame approach discretizes the Miura-ori unit cell, showing the distribution of truss elements along the creases and across the facets, as well as the nodal masses.  (d) Detailed kinematics and mechanics set up to analyze the bending and stretching along the truss \#$pq$. Notice that $\mathbf{m}^{(j)}$ and $\mathbf{n}^{(j)}$ are the current surface normal vectors defined by triangles \#$pqr$ and \#$pqv$, respectively. (e) The bending of the Miura-ori sheet under its weight. This simulation serves to validate appropriate material property assignments.}
	\vspace{0.1in}
	\label{fig:truss}
\end{figure}

To investigate this origami reservoir's computing capacity, one must first obtain its time responses under dynamic excitation.  To this end, we adopt and expand the lattice framework approach to simulate its nonlinear dynamics \cite{Schenk2011, Liu2017, Ghassaei2018}.  In this approach, origami creases are represented by pin-jointed stretchable truss elements with prescribed spring coefficient $K_{s}$.  Folding (or bending) along the crease line is simulated by assigning torsional spring coefficient $K_{b}$ (Figure \ref{fig:truss} (b)).  We further triangulate the quadrilateral facets with additional truss elements to estimate the facet bending with additional torsional stiffness (typically, $K_{b}$ across the facets is larger than those along the creases). Therefore, this approach discretizes the continuous origami sheet into a network of pin-jointed truss elements connected at the nodes.  A typical reservoir consists of an interconnected network of units governed by nonlinear dynamics, and the origami reservoir, in this case, consists of a network of nodes with their interconnections defined by the underlying crease pattern.  The corresponding governing equations of motion, in terms of node \#p's displacement ($\mathbf{x}_p$) as an example, are:
\begin{equation}
	m_p\ddot{\mathbf{x}}_p^{(j)}=\mathbf{F}_{p}^{(j)},
\end{equation}
where the superscript ``$(j)$'' represents the $j^\text{th}$ time step in numerical simulation, and $m_p$ is the equivalent nodal mass.  Unless noted otherwise, the mass of the origami sheet is assumed to be equally distributed to all its nodes. $\mathbf{F}_{p}^{(j)}$ is the summation of internal and external forces acting on this node in that
\begin{equation}
	\mathbf{F}_{p}^{(j)} = \sum \mathbf{F}_{s,p}^{(j)} + \sum \mathbf{F}_{b,p}^{(j)} + \mathbf{F}_{d,p}^{j}+\mathbf{F}_{a,p}^{(j)}+m_p \textbf{g},
\end{equation}
where the five terms on the right hand side are the forces from truss stretching, crease/facet bending, equivalent damping, external actuation, and gravity, respectively. The formulation of these forces are detailed below.

\textbf{Truss stretching forces}: The truss elements are essentially elastic springs with axial stretching stiffness ($K_{s}^{(j)} = EA/l^{(j)}$).  Here, $EA$ is the material constant, and $l^{(j)}$ is the truss element's length at the current $j^\text{th}$ time step.  Thus, the axial stiffness is updated at each time-step, accommodating the truss element's increase in stiffness as it is compressed and vice-a-versa. The stretching forces from a truss connecting node \#p and one of its neighbouring nodes \#$q$ is, 
\begin{equation}
    \mathbf{F}_{s,p}^{(j)} = -K_{s}^{(j)}\left(l_{pq}^{(j)}-l_{pq}^{(0)}\right) \frac{\mathbf{r}_p^{(j)} - \mathbf{r}_q^{(j)}}{|\mathbf{r}_p^{(j)} - \mathbf{r}_q^{(j)}|}
\end{equation}
where $l_{pq}^{(0)}$ is the truss length at its initial resting state. $\mathbf{r}_p^{(j)}$ and $\mathbf{r}_q^{(j)}$ are the current position vectors of these two nodes, respectively. To calculate the total truss stretching forces acting on node \#$p$, similar equations apply to all of its neighbour nodes through trusses (e.g., node $q$, $r$, $s$, $t$, $u$, and $v$ in Figure \ref{fig:truss}(c)).

\textbf{Crease/facet bending forces}:  The crease folding and facet bending are simulated with torsional spring coefficient ($K_{b}^{(j)}=k_b l^{(j)}$), where $k_b$ is torsional stiffness \emph{per unit length}.  Here, we adopt the formulation developed by Liu and Paulino \cite{Liu2017}.   For example, the force acting on nodes \#$p$ due to the crease folding along the truss between \#$p$ and \#$q$ is:
\begin{equation} \label{eq:Fb}
\mathbf{F}_{b,p}^{(j)}= -K_{b}^{(j)}(\varphi_{pq}^{(j)}-\varphi_{pq}^{(0)})\frac{\partial \varphi_{pq}^{(j)}}{\partial \mathbf{r}_p^{(j)}}
\end{equation}
where $\varphi_{pq}^{(j)}$ is the current dihedral angle along truss $pq$ (aka. the dihedral angle between the triangles \#$pqr$ and \#$pqv$ in \ref{fig:truss}(d)), and $\varphi_{pq}^{(0)}$ is the corresponding initial value. $\varphi_{pq}^{(j)}$ can be calculated as
\begin{align} \label{eq:phi1}
    \varphi_{pq}^{(j)} &= \eta \arccos \left(\frac{\mathbf{m}^{(j)} \cdot  \mathbf{n}^{(j)}}{|\mathbf{m}^{(j)}| |\mathbf{n}^{(j)}|}\right) \text{ modulo } 2\pi
\end{align}
\begin{align} \label{eq:phi2}
    \eta &= 
	\begin{cases}
		\text{sign} \left(\mathbf{m}^{(j)} \cdot \mathbf{r}_{pv}^{(j)} \right),& \mathbf{m}^{(j)} \cdot \mathbf{r}_{pv}^{(j)} \neq 0\\
		1.              & \mathbf{m}^{(j)} \cdot \mathbf{r}_{pv}^{(j)} = 0
	\end{cases}
\end{align}

Here, $\mathbf{m}^{(j)}$ and $\mathbf{n}^{(j)}$ are current surface normal vector of the triangles \#$pqr$ and \#$pqv$, respectively, in that $\mathbf{m}^{(j)} = \mathbf{r}_{rq}^{(j)} \times \mathbf{r}_{pq}^{(j)}$ and  $\mathbf{n}^{(j)} = \mathbf{r}_{pq}^{(j)} \times \mathbf{r}_{pv}^{(j)}$. In addition,  $\mathbf{r}_{pq}^{(j)}=\mathbf{r}_p^{(j)}-\mathbf{r}_q^{(j)}$, $\mathbf{r}_{rq}^{(j)}=\mathbf{r}_r^{(j)}-\mathbf{r}_q^{(j)}$, and $\mathbf{r}_{pv}^{(j)}=\mathbf{r}_p^{(j)}-\mathbf{r}_v^{(j)}$. This definition of $\varphi_{pq}^{(j)}$ ensures that the folding angle for valley crease lies in $(0,\pi]$ and the folding angle for mountain crease lies in $(\pi,2\pi]$. The derivative between folding angle $\varphi_{pq}^{(j)}$ and the nodal \#$p$'s current position vector is
\begin{equation}
    \frac{\partial \varphi_{pq}^{(j)}}{\partial \mathbf{r}_p^{(j)}}=\left(\frac{\mathbf{r}_{pv}^{(j)} \cdot \mathbf{r}_{pq}^{(j)}}{|\mathbf{r}_{pq}^{(j)}|^2}-1 \right)\frac{\partial \varphi_{pq}^{(j)}}{\partial \mathbf{r}_v^{(j)}}-\left(\frac{\mathbf{r}_{rq}^{(j)} \cdot \mathbf{r}_{pq}^{(j)}}{|\mathbf{r}_{pq}^{(j)}|^2} \right)\frac{\partial \varphi_{pq}^{(j)}}{\partial \mathbf{r}_r^{(j)}}
\end{equation}
where
\begin{align}
    \frac{\partial \varphi_{pq}^{(j)}}{\partial \mathbf{r}_r^{(j)}}  &=  \frac{|\mathbf{r}_{pq}^{(j)}|}{|\mathbf{m}^{(j)}|^2}\mathbf{m}^{(j)},\\
    \frac{\partial \varphi_{pq}^{(j)}}{\partial \mathbf{r}_v^{(j)}}  &= -\frac{|\mathbf{r}_{pq}^{(j)}|}{|\mathbf{n}^{(j)}|^2}\mathbf{n}^{(j)}.
\end{align}

Again, to calculate the total crease folding and facet bending forces acting on node \#$q$, similar equations apply to trusses connected to this node (e.g., truss $pq$, $pr$, $ps$, $pt$, $pu$, and $pv$ in Figure \ref{fig:truss}(b)).

\textbf{Damping forces}: Estimating damping ratio and damping force is essential to achieve realistic dynamic responses and reduce numerical simulation error accumulation.  In this study, we follow the formulation developed in \cite{Hiller2014, Ghassaei2018}.  This formulation first calculates the average velocity of a node with respect to its neighboring nodes ($\mathbf{v}_\text{avg}^{(j)}$) to effectively remove the rigid body motion components from the relative velocities and ensure that these components are not damped.  Then damping force $\mathbf{F}_{d,p}^{(j)}$ applied on node \#$p$ is given by
\begin{align} 
	\mathbf{F}_{d,p}^{(j)} &= -c_{d}^{(j)}(\mathbf{v}_{p}^{(j)} - \mathbf{v}_\text{avg}^{(j)}) \\
	c_{d}^{(j)} &= 2\zeta\sqrt{K_{s}^{(j)} m_p}
\end{align} 	
where $c_d^{(j)}$ is the equivalent damping coefficient, and $\zeta$ is the damping ratio. 

\textbf{Actuation force}: In the origami reservoir, two types of creases receive actuation.  The first type is ``input creases,'' and they receive input signal $u(t)$ required for emulation and output modulation tasks.  The second type is ``feedback creases,'' and they receive reference or current output signal $z(t)$ required by all computing tasks in this study except for the emulation task (more on the applications of input and feedback creases in Section \ref{sec:PRCsetup}). In the case of multiple outputs, different groups of feedback creases are present. Here, the selection of input and feedback creases are random.  
There are many methods to implement actuation to deliver input $u(t)$ and reference/feedback signal $z(t)$ to the reservoir.  For example, the actuation can take the form of nodal forces on a mass-spring-damper network \cite{Hauser2011, Hauser2012}, motor generated base rotation on octopus-inspired soft arm \cite{Nakajima2013}, or spring resting length changes in a tensegrity structure \cite{Caluwaerts2011}.   In origami, the actuation can take the form of moments that can fold or unfold the selected creases.  We assume that the resting angle $\varphi^{(0)}$ of the input and feedback creases will change --- in response to the actuation at every time step --- to a new equilibrium $\varphi_{a,0}^{(j)}$ in that \cite{Paul2006b, Caluwaerts2011}
\begin{align}
	\varphi_{a,0}^{(j)} & = W_\text{in}\tanh (u^{(j)} )+\varphi^{(0)}	 \quad \text{for input creases;}	\\
    \varphi_{a,0}^{(j)} & = W_\text{fb}\tanh (z^{(j)} )+\varphi^{(0)}	 \quad \text{for feedback creases.}
\end{align}
where $W_\text{in}$  and $W_\text{fb}$ are the input and feedback weight associated with these actuated creases. They are assigned before the training and remain unchanged after that. $u^{(j)}$  and $z^{(j)}$  are the input and feedback signal at the $j^\text{th}$  time step. The magnitude of $W_\text{in}$  and $W_\text{fb}$  are selected such that $ \varphi_{a,0}^{(j)} \in [0,2\pi]$ and consistent with the folding angle assignment. This approach of assigning new equilibrium folding angles is similar to traditional neural network studies that use $\tanh$ as a nonlinear activation function to transform function $z(t)$ into a new one with magnitudes between $[-1,1]$.  Additionally, it prevents actuator saturation due to spurious extreme values of $z(t)$. Denote the torsional stiffness of actuated creases by $K_{b,a}^{(j)}$, and we can update Equation (\ref{eq:Fb}) for the actuated creases (using node \#$p$ as an example)
\begin{align} 
	\mathbf{F}_{a,p}^{(j)} = -K_{b,a}^{(j)}\left(\varphi_{pq}^{(j)}-\varphi_{a,0,pq}^{(j)}\right) \frac{\partial \varphi_{pq}^{(j)}}{\partial \mathbf{r}_p^{(j)}},
\end{align}

The calculation of other terms in this equation are the same as those in the force from crease folding and facet bending. Once the governing equations of motion are formulated, they are solved using MATLAB's \texttt{ode45} solver with $10^{-3}$ second time-steps. Although the governing equation of motions use nodal displacement $\mathbf{x}^{(j)}$ as the independent variables, we use the dihedral crease angles $\varphi^{(j)}$ as the \emph{reservoir state} variables to characterize the origami's time responses.  This is because measuring crease angles is easier to implement by embedded sensors, and $\varphi^{(j)}$ can be directly calculated from $\mathbf{x}^{(j)}$ via the Equations \ref{eq:phi1} and \ref{eq:phi2}.

\subsection{Setting Up Reservoir Computing} \label{sec:PRCsetup}
Similar to the actuated creases (aka. input creases and feedback creases), we designate ``sensor creases'' for measuring the reservoir states.  We denote $N_a$ as the number of actuated creases, and $N_s$ for sensor creases. It is worth noting that, the actuated creases are typically small subset of all origami creases (i.e., $N_a < N$). The sensor creases, on the other hand, can be all of the origami creases ($N_s = N$) or a small subset as well ($N_s<N$).

Once the selections of input, feedback, and sensor creases are completed, one can proceed to the computing. Physical reservoir computing for tasks that require feedback (e.g., pattern generations in Section \ref{sec:pat}, and output modulation in \ref{sec:op}) consists of two phases: The ``training phase'' and ``closed-loop phase.''  While the emulation tasks require the training phase only (Section \ref{sec:emu}).

\begin{figure}[t]
	\centering
	\includegraphics[scale=1.0]{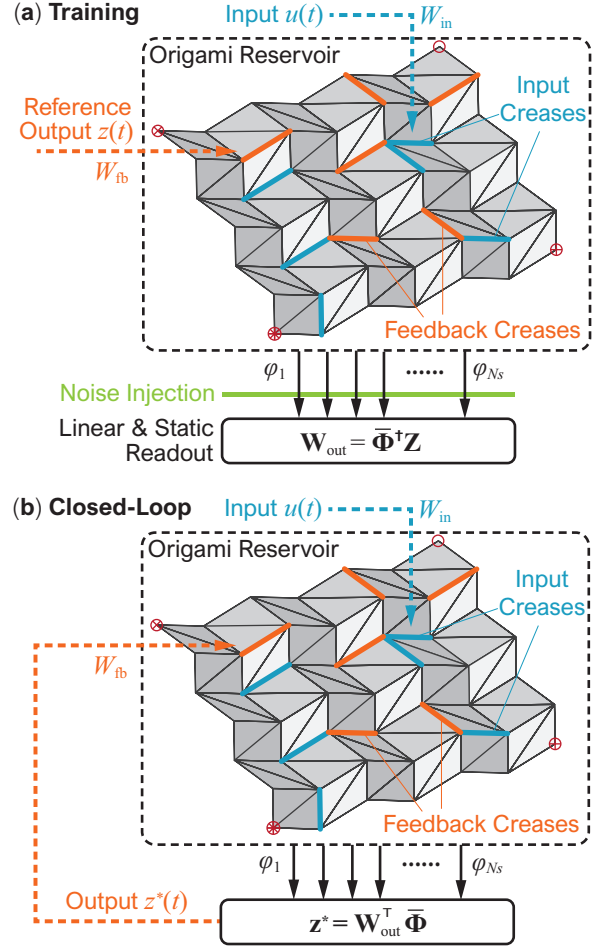}
	\vspace{0.1in}
	\caption{The setup of physical reservoir computing with origami. (a) The training phase.  The feedback creases receive the reference (or targeted) output $z(t)$; while white noise is added to the reservoir state vector $\mathbf{\Phi}(t)$ before calculating output weights $\mathbf{W}_\text{out}$; (b) The closed-loop phase. The output weights obtained in the training phase are used to calculate the current output, which is fed back to the feedback creases.}
	\vspace{0.1in}
	\label{fig:PRC}
\end{figure}

\textbf{Training phase}:  In this phase, we use the teacher forcing to obtain the readout weights $W_i$ corresponding to every reservoir state (aka. the dihedral angles of the sensor creases).  Suppose one wants to train the reservoir to generate a nonlinear time series $z(t)$ (aka. the reference output).  The feedback creases receive the reference output and it dynamically excites the origami reservoir under an open-loop condition without feedback (Figure \ref{fig:PRC}(a)).   The reservoir states $\varphi^{(j)}$ at every time step are measured and then compiled into a matrix $\mathbf{\Phi}$.   

Once the numerical simulation is over, we segregate the reservoir state matrix $\mathbf{\Phi}$ into the washout step, training step, and testing step.  The washout step data is discarded to eliminate the initial transient responses.  We then calculate the output readout weights $W_i$  using the training step data via simple linear regression:
\begin{align}
	   	\mathbf{W}_\text{out}= [\mathbf{1} \; \mathbf{\Phi}]^{+} \mathbf{Z}= \mathbf{\bar{\Phi}} ^{+} \mathbf{Z}
\end{align}
where, $[.]^{+}$ refers to the Moore-Penrose pseudo-inverse to accommodate non-square matrix.  $\mathbf{1}$ is a column of ones for calculating the bias term $W_{\text{out},0}$ to shift the fitted function when necessary.  $\mathbf{Z}$ contains the reference signals at each time step, and it is a matrix if more than one references present.   Lastly, we use testing step data to verify reservoir performance.  It is worth noting that white noise of amplitude $10^{-3}$ is superimposed on the reservoir state matrix during training to ensure the robustness of the readout result against numerical imperfections, external perturbations \cite{Hauser2012}, and instrument noise in ``real-world'' applications. 

\textbf{Closed-loop phase}: Once the training phase is over and readout weights are obtained, we run the reservoir in the closed-loop condition.  That is, instead of using the reference output $z(t)$, the current output $z^*(t)$ is sent to the feedback creases (Figure \ref{fig:PRC}(b)), and 
\begin{align}
		z^*(t) = W_{\text{out},0}+\sum_{i=1}^{N_s} W_{\text{out},i}\varphi_i(t)= \mathbf{W}_{out}^T \bar{\mathbf{\Phi}}
\end{align}
where, $N_s$ is the number of sensor creases, and $\bar{\mathbf{\Phi}}= [\mathbf{1} \; \mathbf{\Phi}]$.  Thus, the reservoir runs autonomously in the closed-loop phase without any external interventions.  

We study the closed loop performance of reservoir by calculating the Mean Squared Error (MSE) using M time-steps as follows:

\begin{equation}\label{Eq:MSE}
   	\text{MSE} = \frac{1}{M} \sum_{j=1}^{M} \left( z(j)-z^*(j) \right)^2
\end{equation}
	
To estimate performance when multiple reference outputs are present, we combine the MSEs by taking a norm over the individual MSEs.

\section{Computation Tasks By the Origami Reservoir} \label{sec:Res}
In this section, we use the origami reservoir to emulate multiple non-linear filters simultaneously, perform pattern generation, and modulate outputs. The baseline variables for the origami geometric design, material properties, and reservoir parameters are given in Table \ref{tab:Ori_res}.

\begin{figure}[t]
	\centering
	\includegraphics[width=1.0\textwidth]{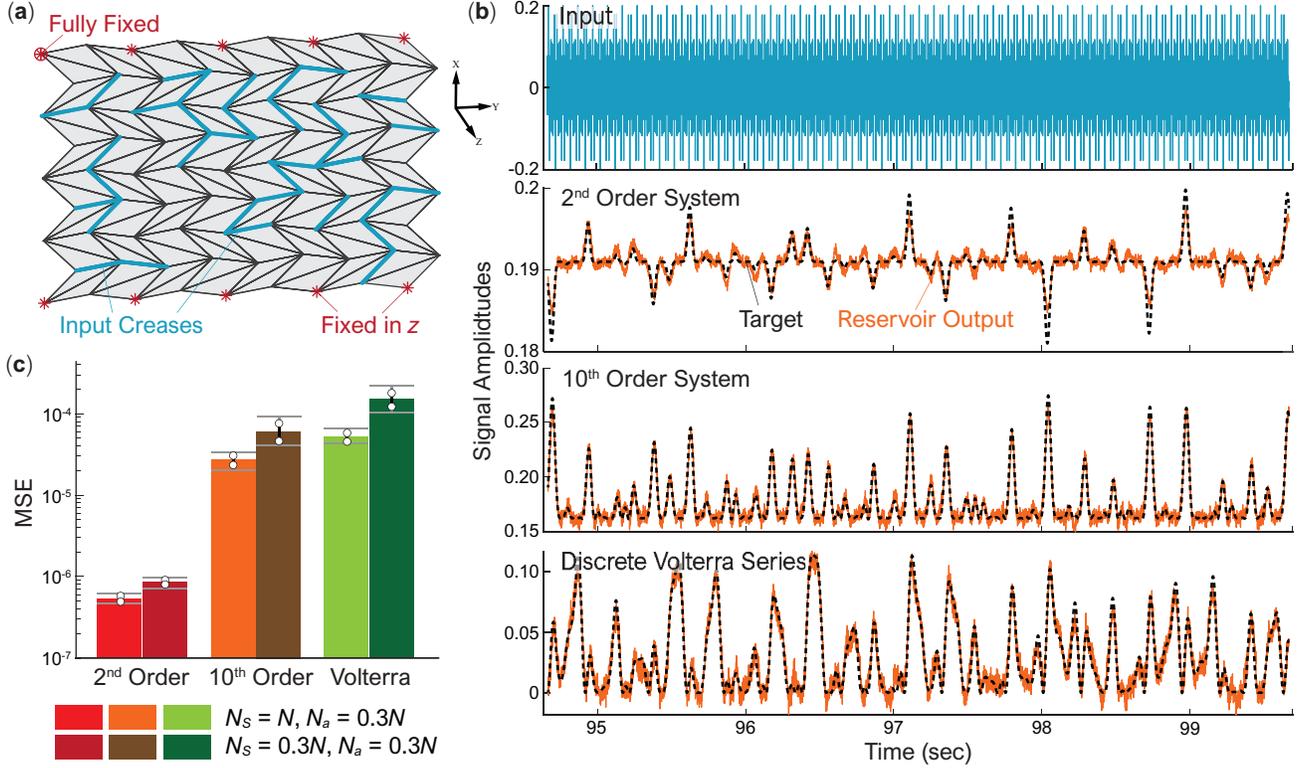}
	\vspace{0.1in}
	\caption{Emulation tasks with the origami reservoir. (a) The Miura-ori reservoir used for this task with input creases highlighted.  Appropriate boundary conditions are also necessary. (b) Examples of trajectories generated in the emulation task including (from top to bottom) input signal $u(t)$, 2$^{nd}$ order, 10$^{th}$ order system, and Volterra series.  Dashed curves are the targeted trajectories, and solid curves are the result of the reservoir.  (c) Error analysis of the emulation tasks. Circles are the standard deviation of MSE, and horizontal bars are the corresponding extreme values.}
	\vspace{0.1in}
	\label{fig:Emulation}
\end{figure}

\begin{table}[t]
\caption{Design of a baseline origami reservoir in this study}
\vspace{0.1in}
\centering
%\footnotesize
\scalebox{0.9}{
\begin{tabular}{p{4cm}|p{2.8cm}}

	\toprule
	\multicolumn{2}{c}{Reservoir size and material properties} \\
	\cmidrule(r){1-2}
	Parameter       & Value \\
	\midrule
	Size            &  9$\times$9  \\
    Nodal Mass      &  7 g  \\
	$ k_s $         & 100 N/m\\
	$ k_c^a $       & 1 N/(m-rad) \\
	$k_c$           & 0.2525 N/(m-rad) \\
   	$ K_f $         & 10 N/(m-rad) \\
   	$\zeta$         & 0.2 \\

	\cmidrule(r){1-2}
	\multicolumn{2}{c}{Geometric design of Miura-ori} \\
	\cmidrule(r){1-2}
	Parameter       & Value \\
	\midrule
	$a$             &  16 mm  \\
	$b$             &  10 mm  \\
	$\gamma$        & 48$\degree$  \\
	$\theta$        & 60$\degree$  \\
			
	\cmidrule(r){1-2}
	\multicolumn{2}{c}{Actuator and sensor creases} \\
	\cmidrule(r){1-2}
	Parameter       & Value  \\
	\cmidrule(r){1-2}
	No. of sensors ($ N_s $)        & $N$  \\
	No. of actuators ($ N_a $)      & 0.45$N$  \\
	No. of Feedback creases         & 0.3$N$  \\
	No. of Input creases            & 0.15$N$ \\
	\bottomrule
	
\end{tabular}
}
\label{tab:Ori_res}
\end{table}

\begin{table*}[h]
\caption{Emulation task functions}
% \footnotesize	
\centering
\scalebox{0.9}{
\begin{tabular}{l|l}
    
	\toprule
	Type  & Functions in discretized form (at $j^{th}$ time step) \\
	\midrule
	Input & $u(j) = 0.2\sin(2\pi f_1 j\Delta t)\sin(2\pi f_2 j \Delta t)\sin(2\pi f_3 j \Delta t)$ \\
	& $ f_1 = 2.11$ Hz, $f_2 = 3.73$ Hz, $f_3 = 4.33$ Hz \\
	\midrule
	$2^\text{nd}$ order system & $z_1(j+1)= 0.4z_1(j)+0.4z_1(j)z_1(j-1)+0.6(u(j\Delta t))^3+0.1  $  \\  
	\midrule
	$10^\text{th}$-order system & $ \displaystyle z_2(j+1)= 0.3z_2(j-1)+0.05z_2(j-1)\sum_{i=1}^{10}z_2(j-i)$ \\
	& $+1.5u((j-10)\Delta t)u((j-1)\Delta t)+0.1   $ \\
	\midrule
	Discrete Volterra series & $ \displaystyle z_3(j+1) = 100\sum_{\tau_1=0}^{T}\sum_{\tau_2=0}^{T}h(\tau_1,\tau_2)u(j-\tau_1)u(nj-\tau_2)$\\
	& $ \displaystyle h(\tau_1,\tau_2) = \exp \left(\frac{(\tau_1\Delta t-\mu_1)^2}{2\sigma_1^2}+\frac{(\tau_2\Delta t-\mu_2)^2}{2\sigma_2^2}\right) $\\
	& $ \mu_1=\mu_2 = 0.1, \sigma_1 = \sigma_2 = 0.05, \Delta t = 10^{-3} $\\
	\bottomrule		
\end{tabular}
}
\label{tab:Emu}
\end{table*}

\subsection{Emulation Task }
\label{sec:emu}

This sub-section shows that the origami reservoir can emulate multiple nonlinear filters simultaneously using a single input.  Such emulation is a benchmark task for evaluating the performance in RNN training \cite{Atiya2000} and prove the multi-tasking capability of physical reservoirs \cite{Hauser2011, Nakajima2013}.  Note that the emulation task involves only the training phase, so there are no feedback creases in this case.  Consequently, we excite the reservoir by sending the input function $u(t)$ to the input creases and train it to find three sets of readout weights in parallel via linear regression.   Here, $u(t)$ is a product of three sinusoidal functions with different frequencies, and the three target non-linear filters are $2^\text{nd}$-order non-linear dynamic system $z_1(t)$, a $10^\text{th}$-order non-linear dynamic system $z_2(t)$, and discrete Volterra series $z_3(n) $ (detailed in Table \ref{tab:Emu}). 

We use a $9\times9$ Miura-ori reservoir in this task, exciting the reservoir from complete rest and training it for 100 seconds.  We discard the first 50 seconds of data as the washout step, use the data from the next 45 seconds to calculate the optimum static readout weights, and then use the last 5 seconds of data to calculate the MSE for performance assessments.  Results in Figure \ref{fig:Emulation} show that the origami reservoir can emulate these three nonlinear filters. As the nonlinearity and complexity of the nonlinear filter increases, MSE also increases (Figure \ref{fig:Emulation}(b)). 

Moreover, we compare the emulation performance when all $N$ creases are used as sensor creases versus when only actuated creases are used as sensors ($N_s = N_a = pN$).  The increase in MSE is marginal in the latter case.  Therefore, the origami satisfies the previously mentioned nonlinearity and fading memory requirements to be a physical reservoir, and one only needs to use the input creases angles as the reservoir states to simplify the reservoir setup.

\subsection{Pattern Generation Task} \label{sec:pat}

Pattern generation tasks are essential for achieving periodic activities such as robotic locomotion gait generation and manipulator control where persistent memory is required.  That is, by embedding these patterns (or limit cycles) in the origami reservoir, one can generate periodic trajectories in the closed-loop.  We again use a $9 \times 9$ Miura-ori reservoir and randomly select $30\%$ of its creases as the feedback creases (this task does not require input creases).  These feedback creases are divided into two groups for the two components of 2D trajectories.  We run the training phase for 100 seconds for each pattern, discard the initial 15 seconds of data as the washout step and use the next 51 seconds' data to calculate the optimum output readout weights. 

\noindent \textbf{Generating non-linear Limit cycles:}
In the following results, the origami reservoir demonstrates its computation capability via generating quadratic limit cycles (LC), Van der Pol limit cycles, and the Lissajous curve in closed-loop. The quadratic limit cycle is defined by two differential equations:
\begin{align} \label{eq:Quad}
	\dot{x}_1  &= x_1 + x_2 - \epsilon(t) x_1 \left(x_1^2 + x_2^2 \right), \\
	\dot{x}_2  &= -2x_1+x_2-x_2 \left( x_1^2 + x_2^2 \right), \label{eq:Quad2}
\end{align} 
where the parameter $\epsilon(t)$ determines the shape of the limit cycle ($\epsilon(t)=1$ in this case). The Van der Pol limit cycle is defined by:
\begin{align}
	\dot{x}_1 &= x_2, \\
	\dot{x}_2 &= -x_1 + \left(1-x_1^2 \right) x_2.
\end{align}

The Lissajous curve is a graph of two sinusoidal signals parameterized by their frequency ratio ($f_1/f_2=0.5$) and phase difference ($\delta=\pi/2$):
\begin{align} 
	x_1 &= \sin\left( f_1 t + \delta \right)\\
	x_2 &= \sin\left( f_2 t \right)
\end{align} 

\begin{figure}[]
	\centering
	\includegraphics[width=1.0\linewidth]{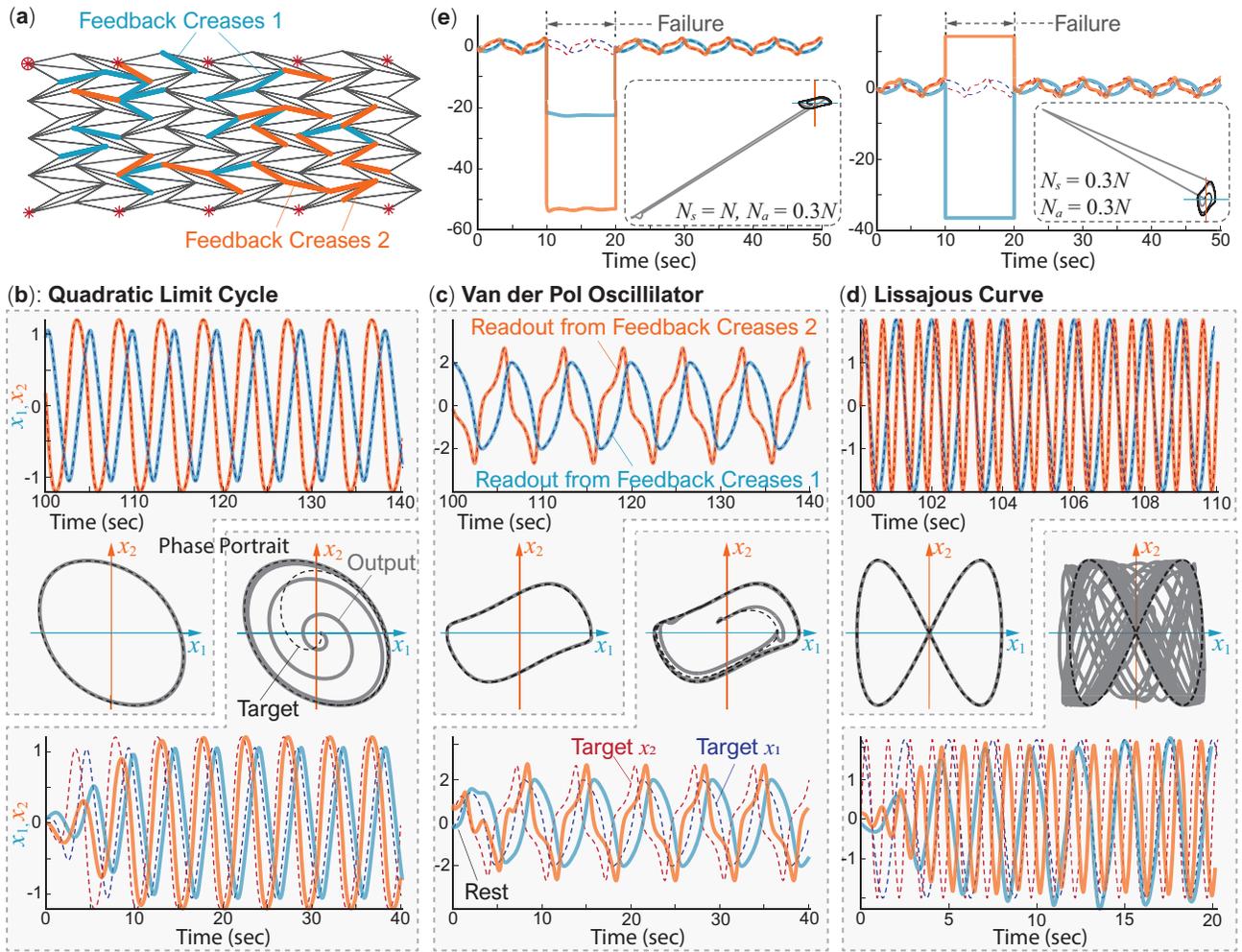}
	\vspace{0.1in}
	\caption{Stable pattern generation under closed-loop using the Miura-ori reservoir.  (a) This task's origami reservoir includes two groups of feedback creases required to generate 2D limit cycles. (b-d) The closed-loop trajectories of quadratic limit cycle, Van der Pol oscillator, and the Lissajous curve, respectively.  In these plots, the first row of time responses shows the closed-loop output after 100s of training.  The third row of time responses shows how the trained reservoir can recover the targeted limit cycles from an initial resting condition.  The corresponding phase portraits are as shown in the second row.  Here, the dashed curves are targeted trajectories, and the solid curves are the reservoir's outputs.   (e) Van der Pol limit cycle recovery after the temporary failure of sensor and actuator creases.  The two simulations are the same except for the number of sensor creases ($N_s=N$ for the first test, $N_s=0.3N$ for the second). The insert figures show the corresponding phase-portraits.}
	\vspace{0.1in}
	\label{fig:lim_cy}
\end{figure}

As shown in Figure \ref{fig:lim_cy}(b), the origami reservoir can generate all three periodic trajectories just by changing the output readout weights.  The MSE for Quadratic LC, Van der Pol LC, and Lissajous curves, calculated using the data for first 10 seconds' closed-loop run (M = 10000), are $3.28 \times 10^{-7}$, $2.03 \times 10^{-5}$, and $5.5 \times 10^{-4}$, respectively. As expected, MSE increases as the complexity of the curve increases.

\textbf{Stability and robustness of the pattern generation:}  
After finding the readout weights, we test the stability of these three limit cycles by starting the origami reservoir from total rest in the close-loop and running it for more than 1000 seconds.  The limit cycle is stable if and only it can recover the pattern from zero initial conditions and stays on target for at least 1000 seconds of simulation  \cite{Hauser2012, Nakajima2013}.  The results in  Figure \ref{fig:lim_cy}(c) indicates that the torsional moments generated from the feedback signals on the feedback creases are sufficient to recover and maintain the three limit cycles from total rest.  Small phase differences occur between generated trajectories and the targets because the reservoir takes a slightly different path than the target, and the Lissajous curve takes more than 15 seconds to recover fully.   Nonetheless, the origami reservoir successfully passes this test.  

To further analyze the robustness of reservoir-generated limit cycles, we simulate actuator and sensor failures.  As the origami reservoir generates the Van der Pol limit cycles in these tests, all feedback and sensor creases stop working (aka. their signals set to zero) for 10 seconds.  We conduct these tests when all creases are used as sensor creases ($N_s=N$) and when only feedback creases are sensor creases ($N_s=N_a=0.3N$).  The simulation results in Figure \ref{fig:lim_cy}(e) show that, although the reservoir diverges to a trajectory far away from the target during the actuator and sensor failure, it can immediately recover the Van der Pol limit cycles after the end of these failures. 

\begin{figure}[h]
	\centering
	\includegraphics[scale=1.0]{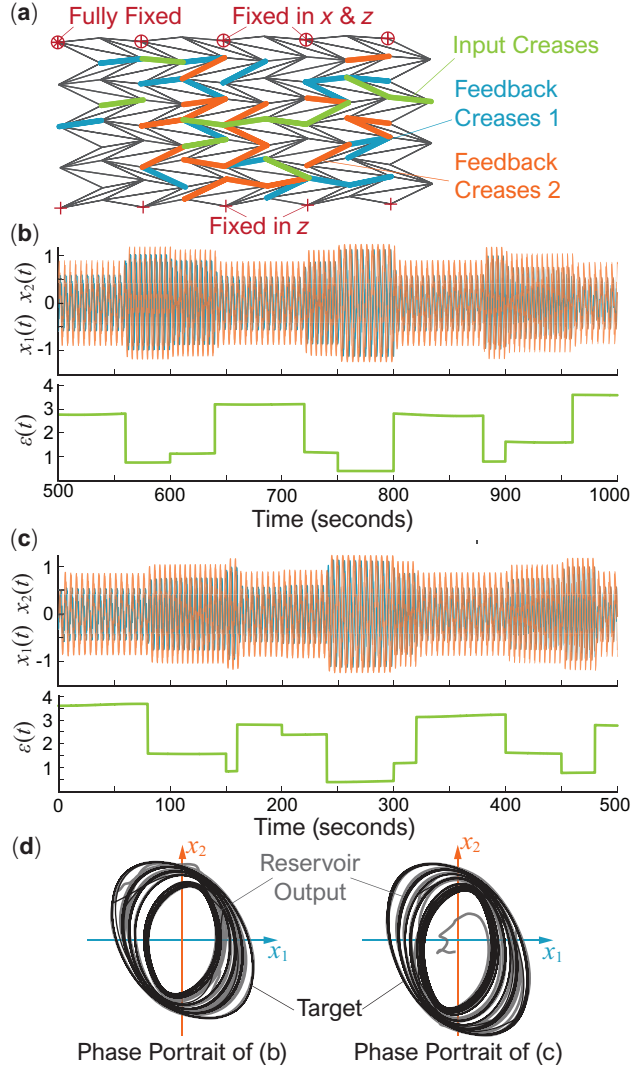}
	\vspace{0.1in}
	\caption{Results of modulation task under closed-loop using The Miura-ori reservoir.  (a) This task's origami reservoir includes two groups of feedback creases and input creases. (b) Quadratic limit cycle trajectories under closed-loop and the corresponding input signal $\epsilon(t)$. The results are obtained after 500 seconds of training. (c) Closed-loop trajectory recovery from the initial resting conditions.  (d) The corresponding phase-portraits, where the targeted trajectories are overlaid on top of the reservoir output.}
	\vspace{0.1in}
	\label{fig:modulation}
\end{figure}

\subsection{Output Modulation Task} \label{sec:op}
Output modulation capability allows the reservoir to adjust its output according to a randomly varying input signal without changing the readout weights.  This ability is also essential for soft robotic control applications because it allows the robot to switch behaviors according to external stimuli or environmental changes.  In this task, we randomly select input creases, which account for $15\%$ of the total creases, in addition to the feedback creases (Figure \ref{fig:modulation}(a)). Moreover, all creases are used as sensor creases ($N_s=N$).  The simulation results in Figure \ref{fig:modulation}(b, c) shows the generated quadratic limit cycles with modulated input (Equation (\ref{eq:Quad}, \ref{eq:Quad2})).  The origami reservoir can react to the input and modulate the magnitude of the quadratic limit cycles.  The MSE is $3.8 \times 10^{-4}$, which is remarkably small, considering this task's complexity.

\section{Correlating Physical Design and Computing Performance} \label{sec:Para}
In this section, we use the mean squared error (MSE) as the metric to examine the connections between the origami reservoir's design and computing performance.  In particular, This analysis aims to investigate the sensitivity of MSE to different parameter changes and identify the optimal origami designs.   To this end, in-depth parametric analyses are conducted to examine the effect of (1) reservoir size and material properties, (2) crease pattern geometry, and (3) feedback and sensor crease distribution.  We use both Van der Pol and quadratic limit cycle generation tasks to ensure the broad applicability of parametric study results.

\begin{table}[]
	\caption{Variables for reservoir size and material properties parametric study}
	\vspace{0.1in}
	% \footnotesize
	\centering
	\scalebox{0.9}{
	
	\begin{tabular}{c|c|c}
		\toprule
		Parameter  &  Base value  &  Distribution \\
		\midrule
		Nodal mass (g) & 7  & [1,50] \\  
		\midrule
		Geometric  & Standard  & \( \sigma = \chi \exp(\frac{-||(N_i-N_j)||}{l}) \)\\ 
		imperfections  &  Miura-ori   & $ \mu = 0 $, $ \chi = 0.4a $, $ l = 4a $ \\
		\midrule
	    Truss torsional  & $ K_b^a = 1 $, & $ K_b^a = 1 $, \\
		stiffness N/(m-rad)  &  $ K_b = 0.2525 $ &  $K_b \in [0.005,0.5] $\\
		\bottomrule		
	\end{tabular}
	}
	\label{tab:Para}
	\end{table}	

\subsection{Reservoir Size, Material Properties, and Vertices Perturbation}	

\begin{figure}[t]
	\centering
	\includegraphics[scale=1.0]{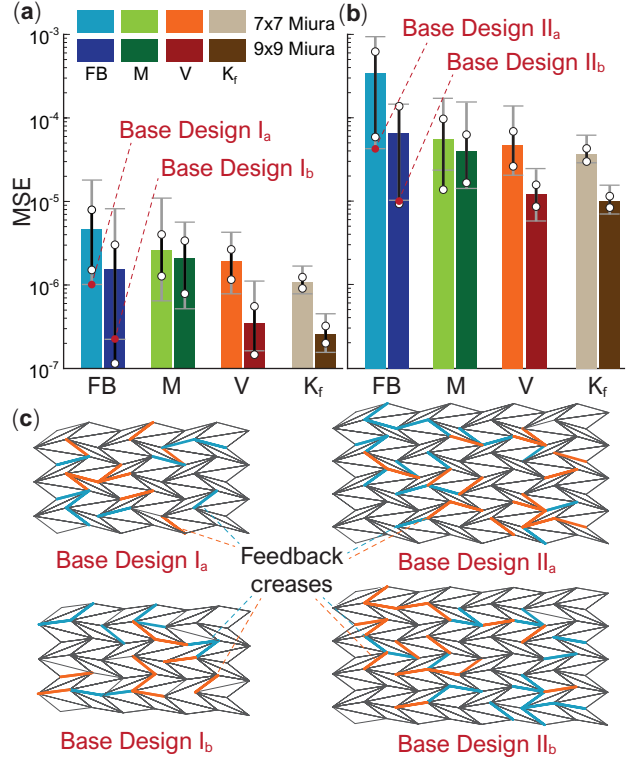}
	\vspace{0.1in}
	\caption{Effect of reservoir size and material properties on the reservoir computing performance. (a) The distribution of MSE from the Quadratic limit cycle simulations using random feedback crease distributions and different design parameter distributions.  Here ``FB'' stands for feedback crease distribution, ``M'' stands for nodal mass distribution, ``V'' stands for origami vertices geometry perturbation, and ``$K_f$'' stands for crease torsional stiffness distribution.  It is worth emphasizing that the ``FB'' results come from one parametric study of 72 unique designs, and the ``M,'' ``V,'' and ``$K_f$'' are results of the subsequent simulation. The bar charts depict the average value, standard deviation (circles), and extreme values (horizontal bars) of MSE. (b) A similar result from the Van der Pol limit cycle generation task.  (c) The feedback crease distributions of the four different baseline designs used in this parametric study.}
	\vspace{0.1in}
	\label{fig:Para_1}
\end{figure}

We observe that feedback crease distribution affects reservoir computing performance quite significantly.  In particular, poorly distributed feedback creases might result in failed pattern generating tasks.  Therefore, we first conduct numerical simulations by randomly changing the feedback crease distributions (72 unique designs in total) and identifying the best performing one (with the least MSE).  We refer to this best performing feedback crease distribution as the \emph{base design} (Figure \ref{fig:Para_1}(a, c)) for the following parametric studies.  Then, we conduct another parametric study regarding the nodal mass, crease stiffness, and vertices perturbation.  We vary these three parameters, one at a time, for 72 randomly selected designs (six batches of jobs in parallel on a computer with 12 cores).  The baseline values and range of the parameters are in Table \ref{tab:Para}.

The origami reservoir performance turns out to be highly sensitive to the nodal mass variation.   As opposed to the uniform nodal mass in base design, a randomly distributed nodal mass can significantly increase or decrease the MSE for both pattern generation tasks.  However, randomly distributing mass in an origami sheet is quite challenging in practical applications. So the use of varying mass distribution should be judicially done based on the particular application at hand.  On the other hand, the origami performance is much less sensitive to the crease torsional stiffness.  By randomly changing the stiffness, one can achieve performance at par with the base design.

Moreover, we investigate the effects of random geometric imperfection in the base designs of origami reservoir.  To this end, we adopt the formulation introduced by Liu et al. \cite{Liu2020}, which introduce small perturbations to the nodal positions in folded origami. Such imperfections are inevitable in practice due to various manufacturing defects. It is found that these small imperfections do not worsen the MSE significantly and in fact could reduce the MSE by a moderate degree (Figure \ref{fig:Para_1}(a),(b)).

It is also worth noting that the larger $9 \times 9$ Miura origami reservoir performs better than the smaller one because larger origami contains more folding angles to constitute the reservoir state matrix.  Therefore, the high-dimensionality of a reservoir is desirable to produce smaller MSE.
		
\subsection{Origami Design}

\begin{figure}[t]
	\centering
	\includegraphics[width=\linewidth]{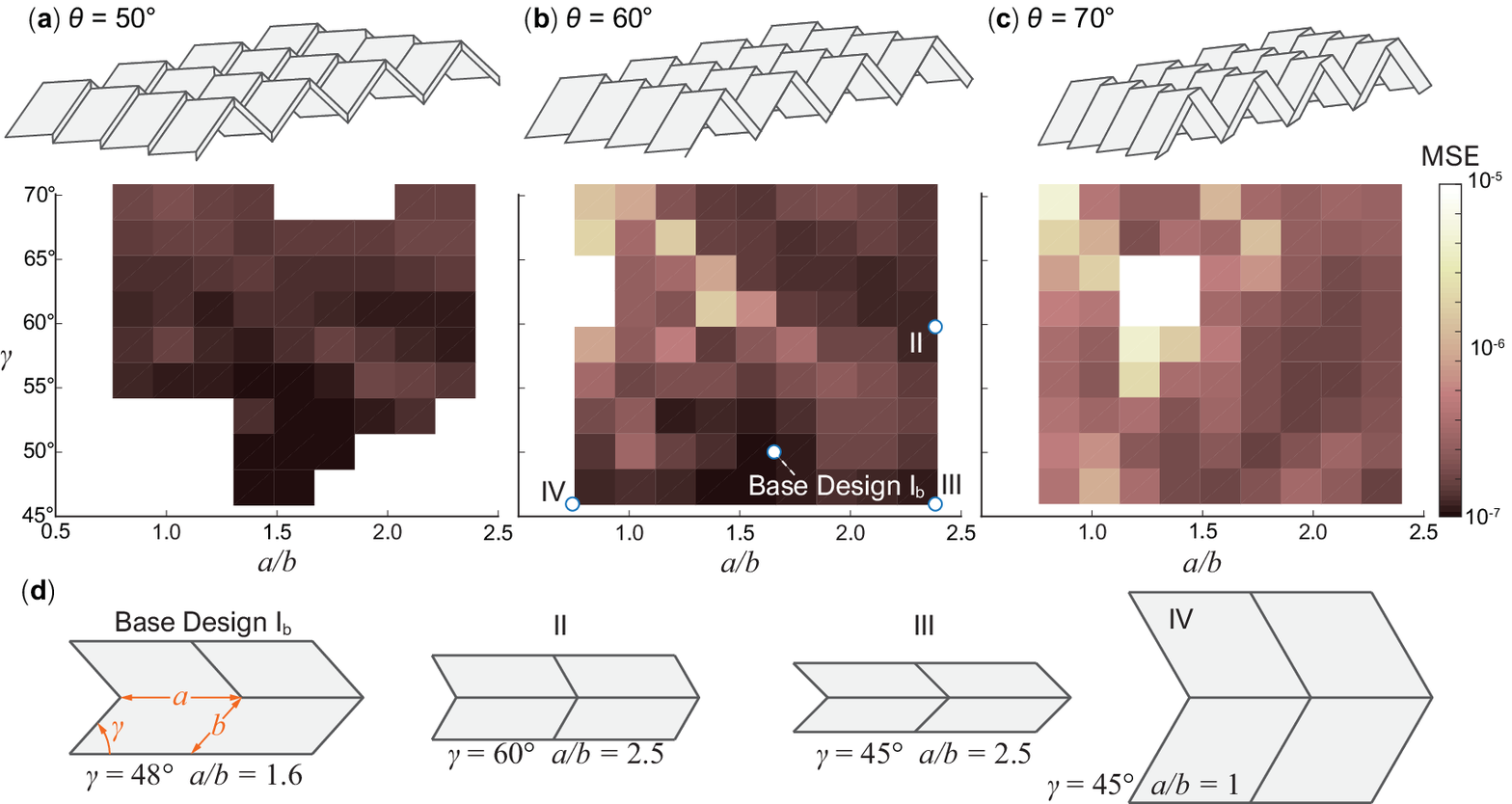}
	\vspace{0.1in}
	\caption{Effect of Miura-ori geometric design on the reservoir performance. (a-c) The Miura-ori geometry and the corresponding landscape of MSE distribution when $\theta=50\degree$, $60\degree$, and $70\degree$, respectively.  The lighter and darker regions correspond to larger and smaller errors, respectively, while the white regions depict origami designs that failed the computing task.  (d) The unit cell geometry of four representative designs with the same crease $a$ length but different sector angles $\gamma$ and crease length ratios $a/b$.}
	\vspace{0.1in}
	\label{fig:Para_2}
\end{figure}

A unique advantage of origami based structures and materials is their considerable freedom to tailor the geometric design.  To this end, we start from the Base Design I of $9 \times 9$ Miura-ori reservoir, vary its crease length ratio $(a/b)$ and internal sector angle $(\gamma)$, and then run the quadratic limit cycle task with 100 crease length and sector angle combinations at three folding angles $(\theta=50\degree, 60\degree, 70\degree )$.   The results of the parametric analysis are shown in Figure \ref{fig:Para_2}.  We observe that, at lower folding angles (flatter origami), the origami reservoir has a higher possibility to fail the pattern generation tasks.  The computing performance improves significantly with a reduced MSE as the origami folds more (or as $\theta$ increases). This trend is probably because  highly folded origami offers an increased range of folding motion.

Moreover, there are two design sets with the lowest MES: $a/b \approx 1.5$, $\gamma \approx 45\degree$, and  $a/b \approx 2.5$,  $\gamma \approx 60\degree$.  Generally speaking, a moderate to high crease-length ratio and small sector angles can create ``skewed'' origami patterns that appear to give better computing performance across all values folding angles.  The best designs here have MSEs at the order of $10^{-7}$, which is of the same magnitude as we found previously by tailoring the nodal mass and crease stiffness.  

\begin{figure}[]
	\centering
	\includegraphics[scale=1.0]{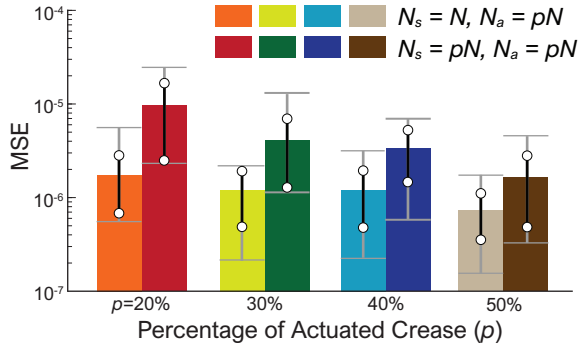}
	\vspace{0.1in}
	\caption{Effect of varying the number of actuator and sensor creases.}
	\vspace{0.1in}
	\label{fig:Para_3}
\end{figure}
	
\subsection{Actuator and Sensors Distribution}
Finally, it is important, for practical applications, to find the minimum amount of input/feedback and sensor creases required for achieving acceptable computing performance.  To this end, we start with the $9 \times 9$ Miura-ori reservoir and conduct two tests.  In the first test, we vary the percentage of feedback creases ($N_a = 0.2N, 0.3N, 0.4N, 0.5N$, each with 24 randomly generated crease distributions) while using all crease dihedral angles to constitute the reservoir state matrix (i.e., $N_s = N$).  In the second test, we use the same feedback crease design and only use these feedback creases' dihedral angles to formulate the reservoir state matrix (i.e., $N_s = N_a$). 

We find that if only $20\%$ of crease are used for feedback, the origami reservoir might fail the quadratic limit cycle task. On the other hand, the MSE reduces only marginally as we increase the percentage of feedback creases beyond $30\%$ (Figure \ref{fig:Para_3}).  Therefore, we can conclude that using only $30\%-40\%$ of total creases as the feedback and sensors crease will provide us an adequate computing performance.  This result is significant because it shows that, even though a large size (high-dimensionality) of the reservoir is essential for computing performance, one does not need to measure (readout) every reservoir state.  In this way, the practical implementation of the origami reservoir can be significantly simplified.

In conclusion, the parametric analyses lay out the strategy to optimize the origami reservoir performance by tailoring the underlying physical and computational design.  A larger origami with a higher-dimension can ensure low computational error, but one only needs to use $30\%$~$40\%$ of its creases as the feedback and sensor creases to tap into the origami's computing capacity.  Meanwhile, the distribution of these feedback and sensor creases must be carefully chosen with extensive simulations.  To further improve computing performance, one can tailor the origami's mass distribution, crease stiffness, and geometric design.  Among these options, optimizing the folding geometry should be the most effective because it is easy to implement in practical applications. 

\section{Application to soft robotic crawling}\label{sec:soro}
\begin{figure}[t]
	\centering
	\includegraphics[width=1.0\linewidth]{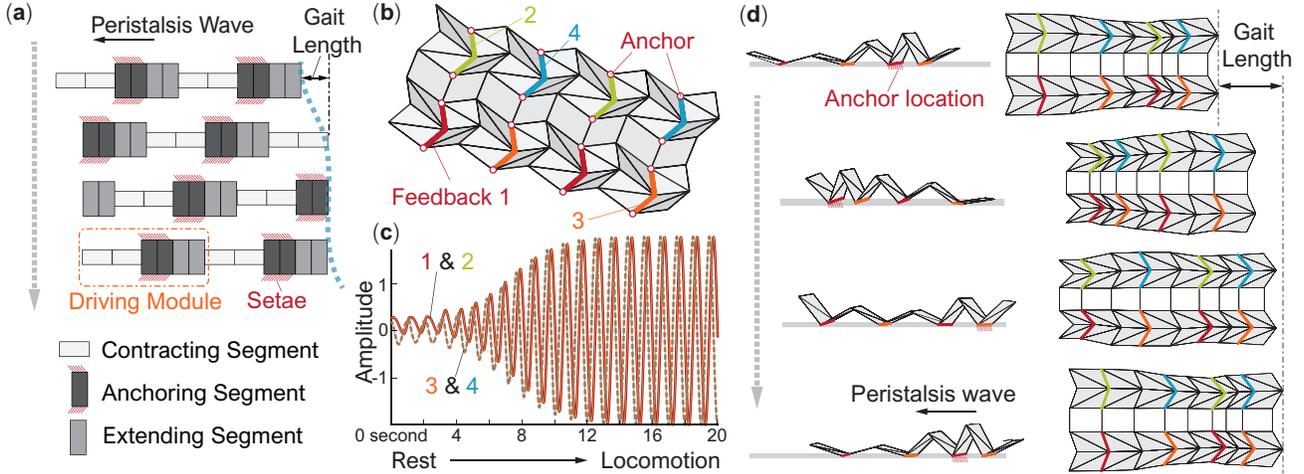}
	\vspace{0.1in}
	\caption{Reservoir computing powered crawling origami robot. (a) The kinematics of a peristaltic locomotion cycle in an earthworm.  For clarity, the earthworm body is simplified and consists of six identical segments organized into two driving modules.  The earthworm body moves forward while the peristaltic wave of anchoring segments (or driving modules) propagates backward.  (b) The design of an earthworm inspired origami crawling robot that features two stripes of Miura-ori connected by a zig-zag shaped ``ridge.'' This robot has four groups of feedback creases.  (c) The closed-loop trajectory generated by the feedback creases after training. (d) Peristaltic locomotion cycle in the origami robot as a result of the generated trajectory.}
	\vspace{0.1in}
	\label{fig:Peristalsis}
\end{figure}

This section demonstrates the application of origami reservoir computing to generate an earthworm-inspired peristaltic crawling gait in a robotic system. The earthworm uses peristalsis to navigate uneven terrain, burrow through soil, and move in confined spaces.  The lack of complex external appendages (aka., legs or wings) makes earthworm-inspired robots ideal for field exploration, disaster relief, or tunnel drilling \cite{Calderon2019,Kamata2018,Fang2017}.  The body of an earthworm consists of segments (metamerism) grouped into multiple ``driving modules'' \cite{Quillin1999, Bhovad2019}. Each driving module includes contracting, anchoring, and extending segments actuated by antagonistic muscles (Figure \ref{fig:Peristalsis}(a)). During peristaltic locomotion, these segments alternately contract, anchor (to the environment with the help of \textit{setae}), and extend to create a propagating peristalsis wave, thus moving the body forward. 

We design an earthworm-inspired origami robot consisting of two $3 \times 9$ Miura-ori reservoir connected via a stiff central bridge (\ref{fig:Peristalsis}(b)). The left and right half of the robots are symmetric in design, and the central bridge design allows differential motion between the two halves to facilitate turning in response to the external input.  In each origami reservoir, we embed two groups of feedback creases (Figure \ref{fig:Peristalsis}(b)) with feedback weights assigned such that their values for the front and back-half are equal but opposite to each other. This arrangement reduces the number of reference outputs needed to generate a crawling gait.   To create a peristalsis locomotion gait, we train the origami reservoirs to generate multiple harmonic signals with a phase difference of $\pi/2$ among them (aka. a pattern generation task shown Figure \ref{fig:Peristalsis}(b)).  We train the robot for 100 seconds and discard the first 15 seconds of data as the washout step.

Also, we apply ideal anchors to the bottom origami creases that are in contact with the surface below.  These anchors are assumed to be kinematically attached to the ground when the associated origami crease folds and relaxed as the crease unfolds (or flattens).  Such anchor design is feasible by leveraging the origami facets' folding motion, as shown in the author's previous study \cite{Bhovad2019}.

Figure \ref{fig:Peristalsis}(d) illustrates the robotic locomotion generated by reservoir computing, while Figure \ref{fig:Peristalsis}(c) depicts the closed-loop response and the limit cycle recovery from total rest (MSE is $3.9 \times 10^{-04})$.  As the origami reservoir generates the multiple harmonic signals with a phase difference, its folding motion naturally ``synchronizes'' to these signals, generating a peristaltic wave of folding and unfolding.  As a result, the robot crawls forward like an earthworm, without using any traditional controllers. 

\section{Summary and Conclusion} \label{sec:end}

We demonstrate the physical reservoir computing capability of origami via extensive benchmark simulations and parametric studies.  First, we develop a simulation environment to study the nonlinear origami dynamics and detail the origami reservoir setup.  This reservoir successfully achieves many computing tasks such as emulation, pattern generation, and modulation, all of which are relevant to robotic applications. We also conduct comprehensive parametric analysis to uncover the linkage between origami reservoir design and its computing performance.  This new knowledge base offers us a guideline to optimize computing performance.  To the authors' best knowledge, this is the first study to rigorously examine the performance of physical reservoir computer from the lens of the physical design.  Finally, we demonstrate how to embed reservoir computing into an origami robot for control without traditional controllers through the example of peristaltic crawling.  

We list four requirements for a mechanical system to be a reservoir in the introduction, and origami satisfies all these requirements.  The tessellated origami structures are inherently high-dimensional.  For example, a $7 \times 7$ Miura-ori with 49 nodes contains $N = 60$ crease dihedral angles, all or a small portion of them can serve as the reservoir states. The nonlinearity of origami partly originates from the nonlinear kinematic relationships between these crease angles and external geometry.  Also, since origami patterns are highly structured (ordered), small perturbations in the material properties, imperfections of crease geometry, and the introduction of local actuation are sufficient to destroy the regularity and create disorder.  These properties make origami highly nonlinear dynamic reservoirs.  The origami reservoir's performance in the emulation task proves that it can act as a nonlinear filter and satisfies fading memory property.  Nonlinear patterns can be embedded into the origami reservoir, and the resulting pattern generation is robust against external disturbances and recoverable under different initial conditions, proving separation property. Finally, adding the feedback can create persistent memory, which is conducive to learning new tasks.

For future robots to work autonomously in unstructured and dynamic environments, the robot body and brain have to work together by continuously exchanging information about the current condition, processing this information, and taking appropriate actions. The physical reservoir computing embodied robots shown in this study presents a step toward this vision.  The reservoir embedded in the robot body directly gathers information from the distributed sensor-actuator network to perform low-level control tasks like locomotion generation.  The resulting soft robot can generate the global target behavior autonomously without controlling every element individually.  Moreover, the generated trajectories could be robust against external disturbances and modulated according to changing working conditions.
 
A challenge in implementing physical reservoir computing is the many sensors and actuators required, even though these sensors and actuators can be simple individually.  Our results contribute in this regard by showing that only a small portion of origami creases are required to be equipped with sensors and actuators to tap into the reservoir computing power.  

In summary, origami reservoir computing provides an attractive pathway for facilitating synergistic collaboration between the soft robot's body and the brain.  The reservoir computing,  coupled with unique mechanical properties that origami can offer --- multi-stability \cite{Li2019, Kamrava2017, Waitukaitis2015}, nonlinear stiffness \cite{Li2019, Schenk2013, Silverberg2014, Kamrava2017}, and negative Poisson's ratio \cite{Li2019, Schenk2013, Kamrava2017} --- opens up new avenues to the next generation of soft robots with embedded mechanical intelligence.

\section*{ACKNOWLEDGMENTS}   

The authors acknowledge the support from the National Science Foundation (CMMI -- 1933124), as well as the Clemson University for the generous allotment of computing time on Palmetto cluster.

% References
\bibliography{report} % bibliography data in report.bib
\bibliographystyle{spiebib} % makes bibtex use spiebib.bst

% \end{multicols}

\end{document}